\documentclass{article}

\usepackage[activate={true,nocompatibility},final,tracking=true,kerning=true,spacing=true,factor=1100,stretch=10,shrink=9]{microtype}
\usepackage[compact]{titlesec}      
\usepackage{wrapfig} 
\newcommand{\squeezeup}{\vspace{-2mm}}

 \usepackage{graphicx}

\usepackage[final]{neurips_2023}

\newcounter{alphasect}
\def\alphainsection{0}

\let\oldsection=\section
\def\section{%
  \ifnum\alphainsection=1%
    \addtocounter{alphasect}{1}
  \fi%
\oldsection}%

\renewcommand\thesection{%
  \ifnum\alphainsection=1%
    \Alph{alphasect}
  \else%
    \arabic{section}
  \fi%
}%

\newenvironment{alphasection}{%
  \ifnum\alphainsection=1%
    \errhelp={Let other blocks end at the beginning of the next block.}
    \errmessage{Nested Alpha section not allowed}
  \fi%
  \setcounter{alphasect}{0}
  \def\alphainsection{1}
}{%
  \setcounter{alphasect}{0}
  \def\alphainsection{0}
}%

\usepackage[utf8]{inputenc} 
\usepackage[T1]{fontenc}    
\usepackage{hyperref}       
\usepackage{url}            
\usepackage{booktabs}       
\usepackage{amsfonts}       
\usepackage{nicefrac}       
\usepackage{microtype}      
\usepackage{xcolor}         

\usepackage{listings}
\usepackage{graphicx}
\usepackage{subfig}
\usepackage{enumitem}
\usepackage{natbib, fancyref}
\usepackage{amsmath,amssymb} 

\usepackage[colorinlistoftodos]{todonotes} 
\usepackage{amsmath,amsfonts}
\usepackage{color}
\usepackage{balance}
\usepackage{xcolor}
\usepackage{multirow}
\usepackage{adjustbox}
\usepackage{soul}
\usepackage{setspace}
\usepackage{eucal}
\usepackage{booktabs}
\usepackage{pifont}
\usepackage{hyperref}
\usepackage{amssymb}
\usepackage{tikz,pgfplots}
\usepackage{soul}

\title{MOFO: MOtion FOcused Self-Supervision
for Video Understanding}

\author{%
  Mona Ahmadian, Frank Guerin, Andrew Gilbert \\
  University of Surrey, UK\\
  \texttt{\{m.ahmadian, f.guerin, a.gilbert\}@surrey.ac.uk} \\
}

\begin{document}

\maketitle

\begin{abstract}
Self-supervised learning (SSL) techniques have recently produced outstanding results in learning visual representations from unlabeled videos. However, despite the importance of motion in supervised learning techniques for action recognition, SSL methods often do not explicitly consider motion information in videos. To address this issue, we propose MOFO (MOtion FOcused), a novel SSL method for focusing representation learning on the motion area of a video for action recognition. MOFO automatically detects motion areas in videos and uses these to guide the self-supervision task. We use a masked autoencoder that randomly masks out a high proportion of the input sequence and forces a specified percentage of the inside of the motion area to be masked and the remainder from outside. We further incorporate motion information into the finetuning step to emphasise motion in the downstream task. We demonstrate that our motion-focused innovations can significantly boost the performance of the currently leading SSL method (VideoMAE) for action recognition. 
Our proposed approach significantly improves the performance of the current SSL method for action recognition, indicating the importance of explicitly encoding motion in SSL.
\end{abstract}

\section{Introduction}

Action recognition is an essential task in video understanding and has been extensively investigated in recent years~\cite{Liu_2022_CVPR,wei2022masked,girdhar2022omnimae}. In video action recognition, supervised deep learning techniques have made significant progress~\cite{tran2015learning,feichtenhofer2019slowfast,lin2019tsm}; However, due to the lack of labels, which must be manually collected, learning to recognise actions from a small number of labelled videos is a difficult task as data collection will be expensive and challenging. It is especially inappropriate for long-tail open vocabulary object distributions across scenes, such as a kitchen. Furthermore, getting annotations for videos is much more difficult due to the large number of frames and the temporal boundaries of when actions begin and end. Therefore, SSL has gained attention due to the problems above.

Supervised methods~\cite{wang2018videos,kwon2020motionsqueeze,patrick2021keeping} have recognised the importance of motion to understand actions because often, key objects are moving in the scene. However, most SSL methods do not explicitly consider motion or use hand-crafted features~\cite{escorcia2022sos}, limiting their effectiveness. In SSL literature, masked autoencoder models~\cite{tong2022videomae} have been proposed to learn the underlying data distribution but without directly emphasising motion autonomously. Even though this model can perform spatiotemporal reasoning over content, the encoder backbone is ineffective in capturing motion representations (we show this later in Fig.~\ref{fig: GradCam}). 
Incorporating motion information is not trivial, especially in egocentric videos. 
Some previous approaches utilized both RGB frames and optical flows ~\cite{han2020self,ni2022motion} to strengthen learning of features but
the primary issue lies in the stability of the results, which can be significantly impacted by camera movement. When the camera moves rapidly, static objects or background pixels exhibit high movement velocities in optical flow. 
Several existing methods leveraged object detection to improve egocentric video recognition~\cite{wang2020symbiotic,wang2020symbiotic,wu2019long,ma2016going}, among which~\cite{wu2019long} also incorporate temporal contexts to help understand the ongoing action. These approaches may have limited uses in real-world systems since they demand time-consuming, labour-intensive object detection annotations and are computationally expensive. In contrast, our framework does not depend on costly object detectors.

\begin{figure}
  \centering
  \includegraphics[width=0.8\linewidth]{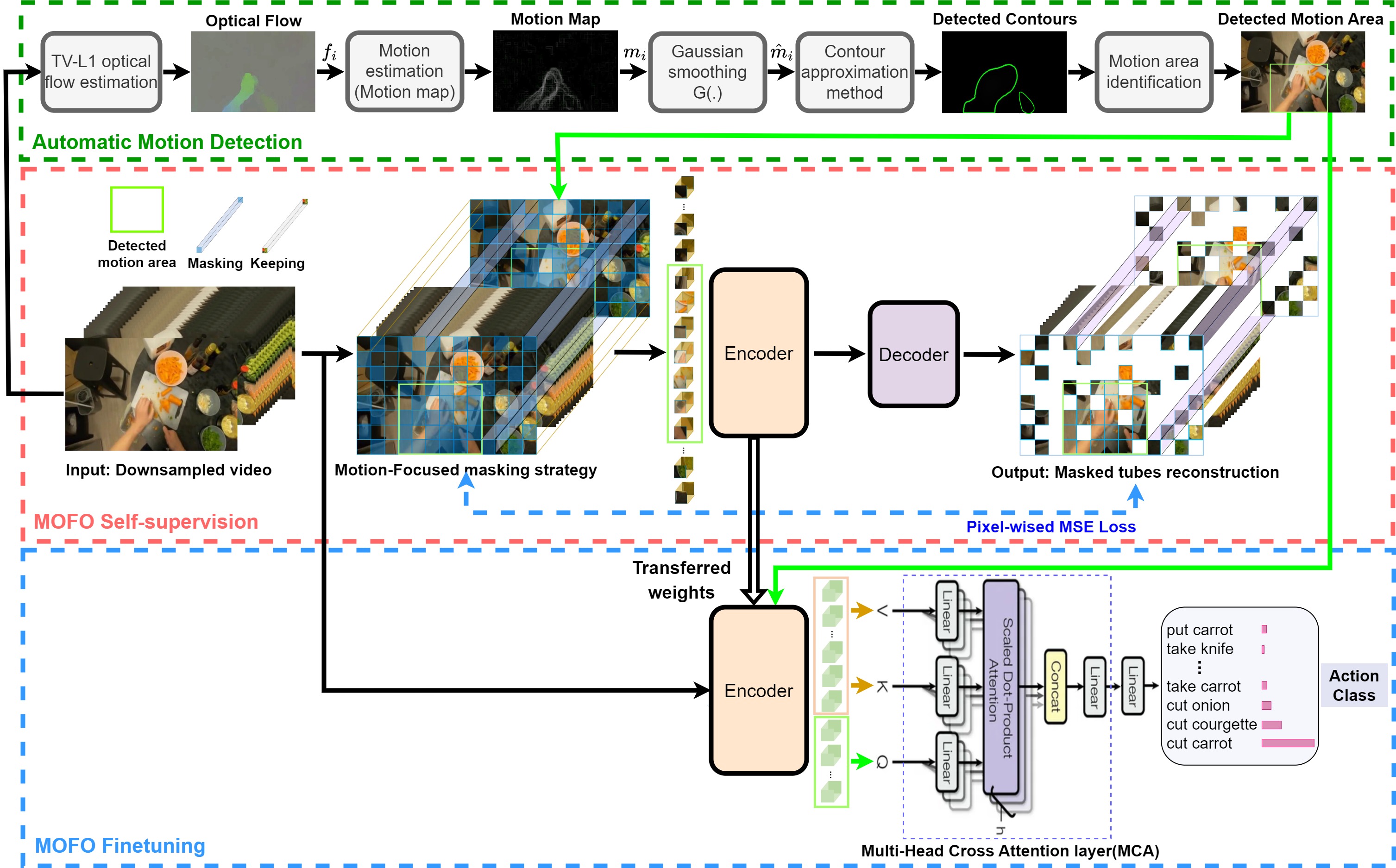}
  \caption{MOFO is a motion-focused self-supervised framework for action recognition.}
  \label{fig:MOFO}
\end{figure}

Fig.~\ref{fig:MOFO} overviews our method,  with three parts: First, our automatic motion area detection using optical flow input to create a motion map to remove camera motion. Second, we propose our new strategy for the SSL pretext task, a reconstruction task focusing more on masking 3D patches on the motion area in the video called MOFO (Motion Focused). Thirdly, the downstream task adaptation step emphasises motion further by integrating motion information during the finetuning training. 
A key contribution of our work is to detect salient objects and motion in the video based on motion boundaries from optical flow. Using the motion boundaries instead of a direct optical flow output mitigates the challenge of camera motion and creates salient areas of movement or interest without a pretrained network. 
Given the motion identification, we suggest extending the self-supervised masking~\cite{tong2022videomae} to include motion understanding. A further contribution is that, during the finetuning stage, MOFO prioritises the motion areas in video data identified as a self-supervision pretext task. Since motion areas contain more information, such as moving objects, actions, and interactions, our proposed model gives them a higher priority by emphasising the masking strategy to be more in the motion area. 


\section{Motion-focused Self-supervised Video Understanding}

\subsection{Automatic motion area detection}\label{sec: motion_detection}
To identify the motion areas without pretrained object detectors, we propose using classical computer vision features, the optical flow vectors. however, these vectors will be affected by camera motion, with static objects or background pixels exhibiting high movement velocities in optical flow when the camera moves rapidly. To mitigate the problem above, we calculate the motion boundaries~\cite{dalal2006human} and use these to define a motion map~\cite{li2021motion}. Therefore, given a video with $T$ frames and a $H \times W$ dimension, we first extract the optical flow vectors representing $ \big\{f_i \in  \mathbb{R}^{H \times W}\big\}_{i=1}^T$ pixel-level motion between two consecutive frames in a video using the TV-L1 algorithm ~\cite{zach2007duality} that offers increased robustness against illumination changes, occlusions, and noise. Then, given the horizontal and vertical displacements of each pixel between the $i$th frame and the $(i + 1)$th frame represented by the flow maps $ u_i, v_i \in \mathbb{R}^{H \times W} $, any
kind of local differential or flow difference cancels out most of the effects of the camera rotation. 
The resulting motion map is defined as:
\begin{equation}
\label{eq:motion map}
m_i= \sqrt{(\frac{\partial{u_i}}{\partial{x}})^{2}+(\frac{\partial{u_i}}{\partial{y}})^{2}+(\frac{\partial{v_i}}{\partial{x}})^{2}+(\frac{\partial{v_i}}{\partial{y}})^{2}}
\end{equation}

where every component denotes the corresponding $x$- and $y$-derivative differential flow frames contributing towards computing $m_i$, representing moving velocity in the $i$-th frame while ignoring the camera motion. As a result, $ m_i \in \mathbb{R}^{H \times W} $ is less influenced by camera motion and considers the moving salients in the $i$-th frame. A low-pass Gaussian filter is used to smooth areas of the image with high-frequency components to further reduce the unwanted noise effect. The Gaussian Smoothing Operator computes an average of the surrounding pixels weighted according to the Gaussian distribution ($G$).


After noise reduction, the next step is to find the boundaries of the motion. To do so, we create contours~\cite{suzuki1985topological}, which are short curves that connect points of the same hue or intensity. We select the two most significant contours in each frame to create a mask that indicates the motion area in a frame of a specific video. The main reason for choosing two contours is that in our datasets, an action is defined by hands and the corresponding object. We create a bounding box around the resulting area that precisely represents the motion in each video. In Fig.~\ref{fig:bbox_ablation}(a), we qualitatively compare our automatic box predictions and the provided supervised annotation for Epic-Kitchens-100 for several sample frames and provide further examples in the Appendix in Fig~\ref{fig:bbx}.

\squeezeup
\subsection{Motion-focused self-supervised learning}\label{sec: mofo_self_supervision}

MOFO uses 3D tube volume embeddings for the self-supervised pretext stage to obtain 3D video patches from frames as inputs. It encodes these with a vanilla ViT~\cite{dosovitskiy2020image} with joint space-time attention as a backbone. We segmented each video into $N$ non-overlapping tubes $\textbf{p}_i \in \mathbb{R}^{H_t \times W_t \times T_t}$. Then, we use a high-ratio tube masking approach to perform masked autoencoder (MAE) pretraining with an asymmetric transformer-based encoder-decoder architecture reconstruction task. Unlike other random masking methods, we explicitly integrate the motion information computed in subsection~\ref{sec: motion_detection} into our masking strategy, resulting in a motion-guided approach to encoding motion for our MAE. Once the motion area is detected, our novel tube masking strategy enforces a mask to be applied on a high portion of the tubes inside the motion area. In other words, a fixed percentage of the tubes (generally 75\%) inside the motion area is always randomly masked to ensure the model is attending more to the motion area at reconstruction time. Therefore, we apply an extremely high masking ratio at random (90\%) while always masking a fixed percentage of the tubes (75\%) inside the motion area. The encoder produces a latent feature representation of the video using input frames with blacked-out regions. The decoder uses the latent feature representation from the encoder. It estimates the missing region using the mean squared error (MSE) loss, computed in pixel space between the masked patches and trained reconstructed outputs. 
Our design encourages the network to capture more useful spatiotemporal structures, making MOFO a more meaningful task and improving the performance of self-supervised pretraining. All models only use the unlabelled data in the training set of each dataset for pertaining. 

\squeezeup

\subsection{Motion-focused finetuning}\label{sec: mofo_finetuning}
Recall that the self-supervised learning protocol is split between a pretraining and finetuning stage. We propose a new approach to focus on the motion area at both the pretext and the finetuning of the model. 
The model is trained end-to-end during finetuning, using the weights of the pretrained network as initialisation for the downstream supervised task dataset.  

As the area inside the motion box has more semantic motion information, we wish to exploit this information for our task by leveraging the detected motion box. On the other hand, the video's setting and any nearby items could provide context for categorising the video clips for the action recognition task. For instance, in the case of washing dishes, the hands can be seen in the sink, but the dishes beside the sink may indicate that the person is washing them. 
Therefore, we propose to use multi-cross attention (MCA)~\cite{nagrani2021attention} in our encoder. MCA is an attention mechanism that mixes two different embedding sequences; the two are from the same modality. Unlike self-attention, where inputs are the same set, during cross-attention, they differ; MCA's main objective is to determine attention scores using data from various information sources. This module resides between the encoder and MLP classifier layers, takes the inner and outer motion box embeddings, and outputs the fused embedding
(see details in Appendix~\ref{Finetune_supp}).

\section{Experiments}
We use two well-known and large datasets to evaluate our proposed approach: \textbf{Something-Something V2 (SSV2)}~\cite{goyal2017something} and \textbf{Epic-Kitchens-100} ~\cite{damen2022rescaling}. Using egocentric videos to predict first-person activity faces many challenges, including a limited field of view, occlusions, and unstable motions, and there is a relative scarcity of labelled data. 

\textbf{Results and analysis}\label{results}
We finetune the learned model for action classification based on our proposed MOFO finetuning approach to evaluate the pretrained model and train on a new downstream task with the learned representation. 
The entire feature encoder and a linear layer are finetuned end-to-end with cross-entropy loss, with recognition accuracy reported in Table~\ref{tab:MOFO pretraining}. We demonstrate significant performance improvement over the other self-supervised approaches, increasing $2.6$\%, $2.1$\%, and $1.3$\% accuracy over the best-performing methods on Epic-Kitchens verb, noun, and action classification and $4.7$\% on Something Something V2 action classification, respectively. In terms of masking ratio, variants are presented in the Appendix, but we found that the 75\% inside masking ratio worked the best.
Our strategy outperforms approaches like OmniMAE~\cite{girdhar2022omnimae}, trained jointly on images and videos by $3.2\%$ in Top-1 accuracy. On Something Something V2, our method outperforms VIMPAC~\cite{tan2021vimpac} and ST-MAE~\cite{NEURIPS2022_e97d1081}, which both use ViT-Large as a backbone, whereas our backbone is vanilla ViT-Base with over $3$x fewer parameters. Compared to VideoMAE~\cite{tong2022videomae}, our approach achieves significantly better results while the number of backbone parameters remains the same. 

\begin{table}[t!]
\centering
\caption{Human activity recognition on \textbf{Epic-Kitchens} and \textbf{Something-Something V2 (SSV2)} in terms of Top-1 and Top-5 accuracy.
}
\vspace{2mm}
\resizebox{1\linewidth}{!}{
  \begin{tabular}{c|c|c|cc|ccc}
    \toprule
    \multirow{3}{*}{Method} & \multirow{3}{*}{Backbone} & \multirow{3}{*}{Param} & \multicolumn{2}{c|}{\textbf{SSV2}} & \multicolumn{3}{c}{\textbf{Epic-Kitchens}} \\
                        
    & & &\multicolumn{2}{c|}{Action} & \multicolumn{1}{c|}{Verb} & \multicolumn{1}{c|}{Noun} & \multicolumn{1}{c}{Action} \\
    
    &  & & Top-1 & Top-5 &\multicolumn{1}{c|}{Top-1} & \multicolumn{1}{c|}{Top-1} & Top-1 \\ 
        \midrule
            
                
        

        
        
   
        
        
   
        
         VIMPAC~\cite{tan2021vimpac} & ViT-L  & 307 & 68.1& - & -  & -  & -\\

         BEVT~\cite{wang2022bevt} & Swin-B  & 88 & 70.6 & - & -  &-  & -\\
        
    
         VideoMAE~\cite{tong2022videomae} & ViT-B  & 87 &70.8 &92.4& 71.6& 66.0  &53.2\\

          ST-MAE~\cite{NEURIPS2022_e97d1081} & ViT-L  & 304 &72.1 & -  & -  & -  &-\\
          
          OmniMAE~\cite{girdhar2022omnimae} & ViT-B  & 87 &69.5 & -& -  & -  & 39.3\\

       Omnivore(Swin-B)~\cite{girdhar2022omnivore} & ViT-B  & - &71.4 & 93.5& 69.5  & 61.7  & 49.9\\

        \midrule
          

          \textbf{MOFO (Proposed)} &ViT-B  & 102& \textbf{75.5} & \textbf{95.3} &  \textbf{74.2}   &  \textbf{68.1} & \textbf{54.5} \\
          \bottomrule
        \end{tabular}
      }
        \label{tab:MOFO pretraining}
\end{table}

\paragraph{\textbf{Visualizing self-supervised representation}}

\begin{wrapfigure}{r}{0.48\textwidth}
\centering
\includegraphics[width=1\linewidth]{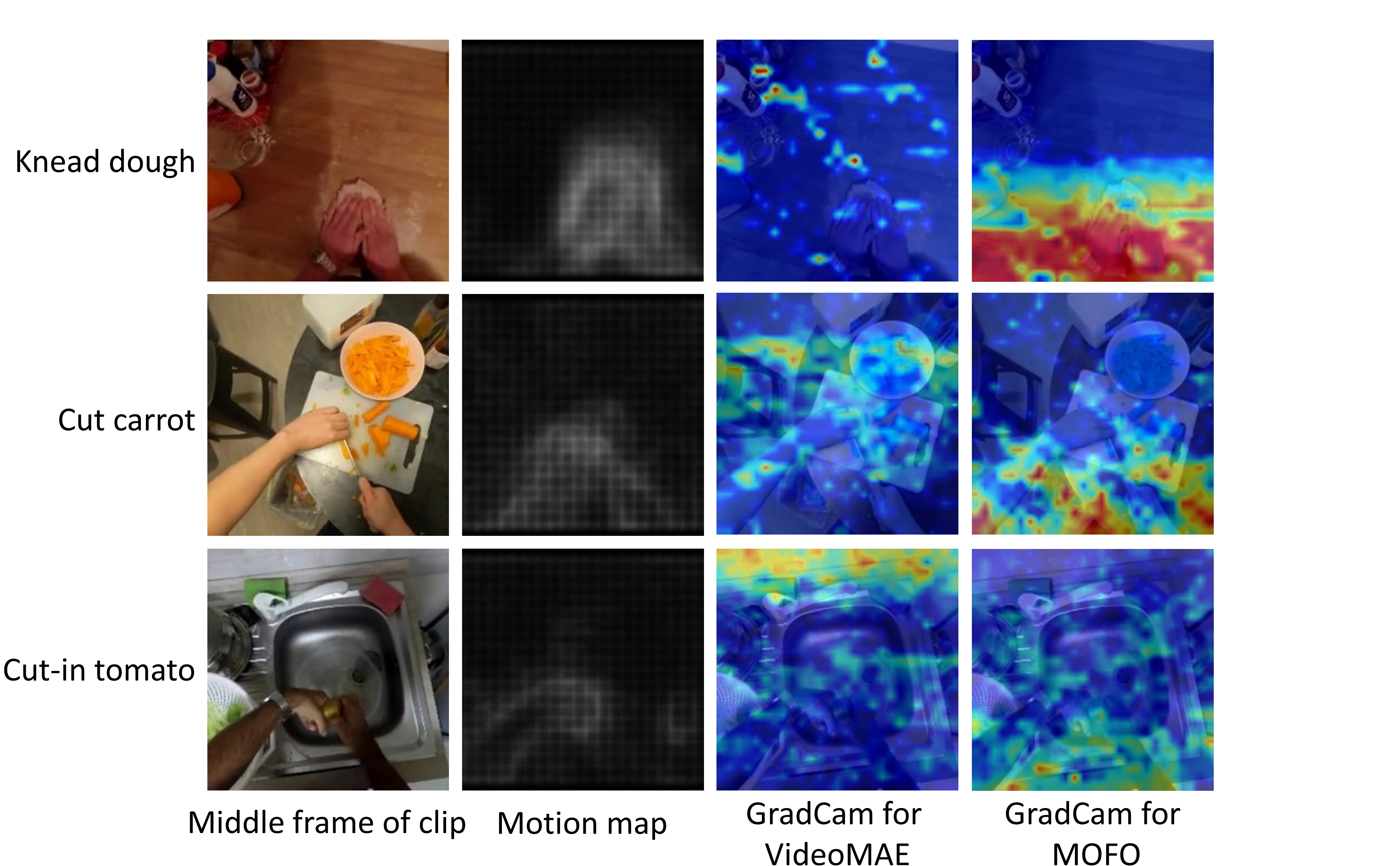}
\caption{Visualisation of the learned features}
\label{fig: GradCam}
\end{wrapfigure}

To further understand the representations learned by MOFO, we utilise GradCAM~\cite{selvaraju2017grad} to create a saliency map highlighting each pixel's importance to show how each pixel contributes to the discrimination of the video clip. Fig.~\ref{fig: GradCam} visualises the middle frame of a video clip, the motion map of the VideoMAE and our MOFO from the fifth attention layer of the ViT-Base backbone. It is interesting to note that for similar actions: \emph{knead dough}, \emph{cut carrot}, and \emph{cut-in tomato}, MOFO is sensitive to the location that is the most significant motion location as detected by our automatic algorithm. 

\section{Conclusion}
MOFO introduces a Motion-Focused technique which explores motion information for enhancing motion-aware self-supervised video action recognition. 
We propose an innovative strategy, an effective self-supervised pretext task, and a modification to masked autoencoding, which focuses masking on the motion area in the video (Motion Focused). 
Extensive experiments on two challenging datasets demonstrate that this context-based SSL technique improves performance in action recognition tasks, and the public code will guide many research directions. 

\paragraph{Acknowledgement}The work was partially funded by a Leverhulme Trust Research Project Grant: RPG-2023-079 "How humans understand video".


\bibliographystyle{plainnat}
\bibliography{egbib.bib}

\section*{Appendix}

We also conducted various ablation studies to examine the design choices made in our proposed strategy. 
\begin{alphasection}
\section{Motion-focused Self-supervised Learning}\label{SSL-supp}
\paragraph{Experimental setting.}
We use the technique of~\cite{denseflow} to extract optical flow from a video to create a motion map, which is 40\% faster by parallelizing IO and computation.
\\MOFO uses ViT-Base as a decoder/encoder backbone, trained for 800 epochs on Something-Something V2 and Epic-Kitchens for the SSL independently. We follow the training and experiential parameters from recent work~\cite{tong2022videomae} to ensure a fair comparison and finetune for 100 epochs with early stopping. The model takes $16$ frames from the video with $224 \times 224$ size and divides the input video into a 3D $16 \times 16\times 8$ patch embeddings, resulting in $H=224$, $W=224$, $T=16$, $H_t=16$, $W_t=16$, $T_t=8$, and $N=392$. While we have a fixed number of input patches for our model, we do not have a fixed number of inner $N_{\text{inner}}$ and outer $N_{\text{outer}}$ embeddings due to varying size of the motion area in each video clip. We report Top-1 accuracy on Epic-Kitchens and Top-1 and Top-5 accuracy on Something-Something V2 on downstream tasks and use Pytorch and DeepSpeed~\cite{li2022deepspeed} on $4$xNVIDIA Quadro RTX-5000 GPU for our experiments.

\paragraph{ Masking ratio.}\label{ratio_supp} VideoMAE~\cite{tong2022videomae} recommended tube masking with an extremely high ratio which helps reduce information leakage during masked modelling. They demonstrated the best efficiency and efficacy with a masking ratio of 90\%. Therefore, we explore the effect of the inside masking ratio for verb classification on Epic-Kitchens in Fig.~\ref{fig:ratio}. It shows that the model pretrained with a masking ratio of 90\% as the general masking ratio for a video and a high ratio for inside masking ratio (75\%) achieves the highest efficiency level. Thus, we continue experimenting with the rest by fixing the inside mask ratio to $75$\%.

\begin{figure}[t!]
    \centering
    \centering
    \begin{tikzpicture}[scale=0.9]
        \begin{axis}[
            title={},
            xlabel={Inside masking ratio (\%)},
            ylabel={Top-1 Accuracy (\%)},
            xmin=65, xmax=100,
            ymin=71, ymax=75,
            xtick={60,65,70,75,80,85,90,95,100},
            ytick={70,71,72,73,74,75},
            legend pos=north west,
            ymajorgrids=true,
            grid style=dashed,
        ]
        \addplot[
            color=blue,
            mark=square,
            nodes near coords,
            ]
            coordinates {
            (70,72.34)(75,72.99)(90,72.82)(95,72.55)
            };
            \legend{Epic-Kitchens}
        \end{axis}
    \end{tikzpicture}
    \caption{The effect of inside masking ratio on Epic-Kitchens-100 dataset for verb classification demonstrates that a high inside masking ratio (75\%) delivers the best efficiency and effectiveness trade-off.}
    \label{fig:ratio}
\end{figure}
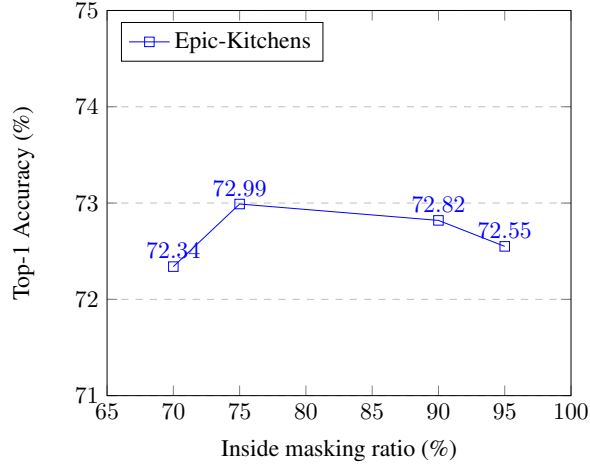

\paragraph{Reconstructed frames}
This section shows several reconstructed image frames from a video in Fig.~\ref{fig:reconstruction_E} and Fig.~\ref{fig:reconstruction_S}.
We use an asymmetric encoder-decoder architecture to accomplish video self-supervised pretraining tube masking with a high ratio for MAE pretraining. We can reconstruct the masked patches using random tube masking by finding the spatially and temporally corresponding unmasked patches in the adjacent frames. The loss function is the mean squared error (MSE) loss between normalised masked tokens and reconstructed tokens in pixel space.
Videos are all randomly chosen from the validation sets of both datasets. Our proposed MOFO model ensures that a fixed number of masks exist within the motion area compared to the VideoMAE model. These examples suggest that, compared to VideoMAE, our MOFO model reconstructs the samples in the motion area significantly more accurately, demonstrating that the model has focused on the motion area. 
We can produce satisfying reconstruction results, mainly when motion occurs with our MOFO, by applying extremely high ratio masking at random (90\%) while always masking a fixed percentage of the tubes (75\%) inside the motion area. 
 
\begin{figure}[htp!]
\centering
\includegraphics[width=1\linewidth]{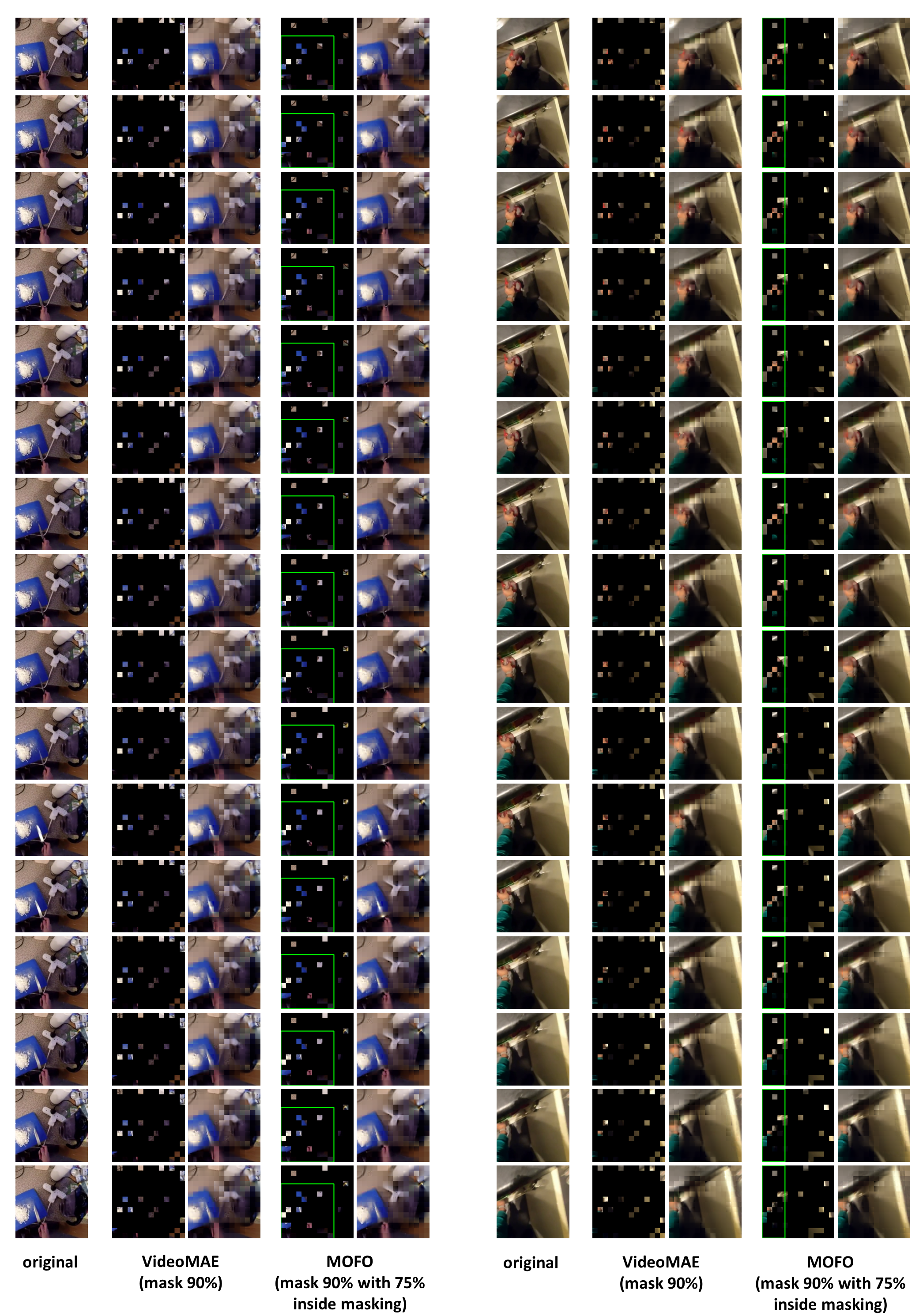}
\caption{Qualitative Comparison on reconstructions using VideoMAE and MOFO on\textbf{ Epic-Kitchens} dataset. MOFO Reconstructions of videos are predicted by MOFO pre-trained with a masking ratio of 90\% and an inside masking ratio of 75\% .}
\label{fig:reconstruction_E}
\end{figure}

\begin{figure}[htp!]
\centering
\includegraphics[width=1\linewidth]{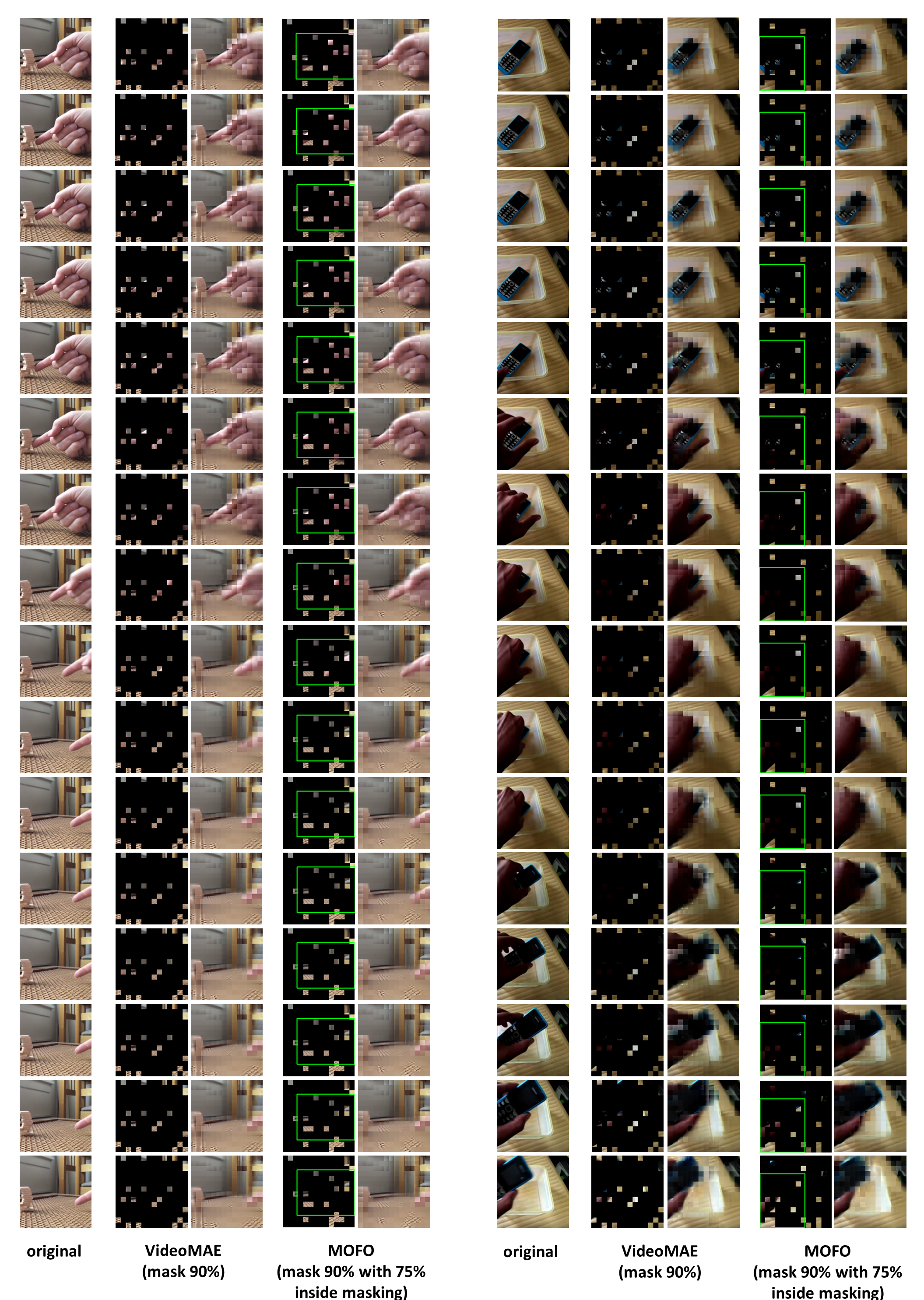}
\caption{Qualitative Comparison on reconstructions using VideoMAE and MOFO on \textbf{Something-Something V2} dataset. MOFO Reconstructions of videos are predicted by MOFO pre-trained with a masking ratio of 90\%  and an inside masking ratio of 75\%.}
\label{fig:reconstruction_S}
\end{figure}

\section{Motion-focused Finetuning}\label{Finetune_supp}
\paragraph{Setup details}\label{MCA}
Given a set of patches $\{\textbf{p}_i\}_1^N$, the transformer yields two sets of embeddings: $\{\textbf{e}^{\text{inner}}\}_{j = 1}^{N_{\text{inner}}}$ for the inner motion boxes and $\{\textbf{e}^{\text{outer}}\}_{k = 1}^{N_{\text{outer}}}$ for the outer ones, as described by:
\begin{equation}
\label{eq:embeddings}
\{\textbf{e}^{\text{inner}}\}_{j = 1}^{N_{\text{inner}}}, \{\textbf{e}^{\text{outer}}\}_{k = 1}^{N_{\text{outer}}} = \text{ViT}\big(\{\textbf{p}_i\}_1^N\big)
\end{equation}
These embeddings are then processed by a cross-attention mechanism, where $Q$, $K$, and $V$ represent query, key, and value, respectively. The CrossAttention function is formalised as follows:
\begin{equation}
\label{eq:cross-attention}
\text{CrossAttention}(Q,K,V) = \text{softmax}\left(\frac{QK^T}{\sqrt{d_k}}\right)V
\end{equation}
where $Q=\textbf{e}^{\text{inner}}, K=V=\textbf{e}^{\text{outer}}$. In the context of multi-head attention, each attention head $i$ is computed by applying the CrossAttention function to the query, key, and value matrices, each weighted by a different learned weight matrix $ W_i^Q \in \mathbb{R}^{d_{model} \times d_q},  W_i^k \in \mathbb{R}^{d_{model} \times d_k}, W_i^V \in \mathbb{R}^{d_{model} \times d_v}$ respectively:
\begin{equation}
\label{eq:head}
\text{head}_i = \text{CrossAttention}(QW_i^Q, KW_i^K, VW_i^V)
\end{equation}
Finally, the fused embedding $\textbf{e}^{\text{fused}}$ is computed by concatenating the results from all attention heads and then applying another learned weight matrix $W^O \in \mathbb{R}^{hd_v\times d_{model}}$. This multi-head cross-attention (MCA) operation can be represented as:
\begin{equation}
\label{eq:fused}
\textbf{e}^{\text{fused}} = \text{MCA}(Q,K,V) = \text{Concat}\left(\text{head}_1, \cdots ,\text{head}_h\right) W^O
\end{equation}
 
We employ $h = 3$ parallel attention layers, or heads, in this work. We also use $d_q = d_k = d_v = d_{model}$ for each.
The model is ultimately finetuned with a cross-entropy loss $\mathcal{L}$ :
\begin{equation}\label{eq:loss}
\begin{gathered}
   \mathcal{L} = -\displaystyle\sum_{n} \mathbf{y}_n \log \mathbf{\hat{y}}_n\\
   \hat{\textbf{y}}= \text{FC}(\textbf{e}^{\text{fused}})
\end{gathered}
\end{equation}

\noindent where, $\mathbf{y}_n$ is the true label for $n$th video clip, $\mathbf{\hat{y}}_n$ is its predicted label, and $\text{FC}$ is the fully connected layers typically used for classification.

\paragraph{MCA hyper-parameters ablation.}\label{MCA_abl}
We list the MCA hyperparameters used in our MOFO finetuning experiments here. We experiment with various head and depth settings when Epic-Kitchens is the target dataset shown in Table~\ref{tab:MOFO_finetune_param}. We experiment with these parameters for the verb task on Epic-Kitchens to find the best choice for the cross-attention layer we suggested for MOFO finetuning. The final head and depth are 3 and 1, respectively.

\begin{table}[htp!]
\centering

    \caption{Ablation experiment for number of head and depth in MOFO finetuning}
\vspace{1mm}
\scalebox{1}{
        \begin{tabular}[t]{c|c|c|c|c}
            \hline
            \multirow{3}{*}{Finetuning method} & \multirow{3}{*}{Backbone training}& \multirow{3}{*}{CA heads} & \multirow{3}{*} {CA depths} & \textbf{Epic-Kitchens} \\
                            
            & & & & Verb  \\
            &  & & & Top-1\\ \hline    
            VideoMAE & VideoMAE  & - & - & $71.6$   \\
           MOFO & VideoMAE & $1$& $1$& $73.5$ \\
            MOFO & VideoMAE &$1$ &$2$ & $73.8$ \\
            MOFO & VideoMAE & $1$& $3$& $73.6$ \\
            MOFO & VideoMAE &$2$ &$1$ & $73.7$ \\
            MOFO & VideoMAE &$2$ & $2$& $73.3$ \\
            MOFO & VideoMAE &$\bf3$ & $\bf1$& $\bf74.0$ \\
            MOFO & VideoMAE &$3$ &$2$ & $73.5$ \\
            MOFO & VideoMAE &$4$ & $1$& $73.8$\\
            MOFO & VideoMAE &$4$ &$2$ & $73.3$ \\
            
        \end{tabular}
        }
        \label{tab:MOFO_finetune_param}
\end{table}

\paragraph{Visualisation of GradCAM using MOFO self-supervision}
We visualise the GradCAM and motion map in Fig.~\ref{fig:CAM} for the samples in which VideoMAE can't identify the class, but our MOFO can. The attention maps show how effective our approach is in capturing the motion area.
Visualisation of important
areas. The heatmap indicates how much the pretrained model
attends to the region.
\begin{figure}[htp!]
\centering
\includegraphics[width=1\linewidth]{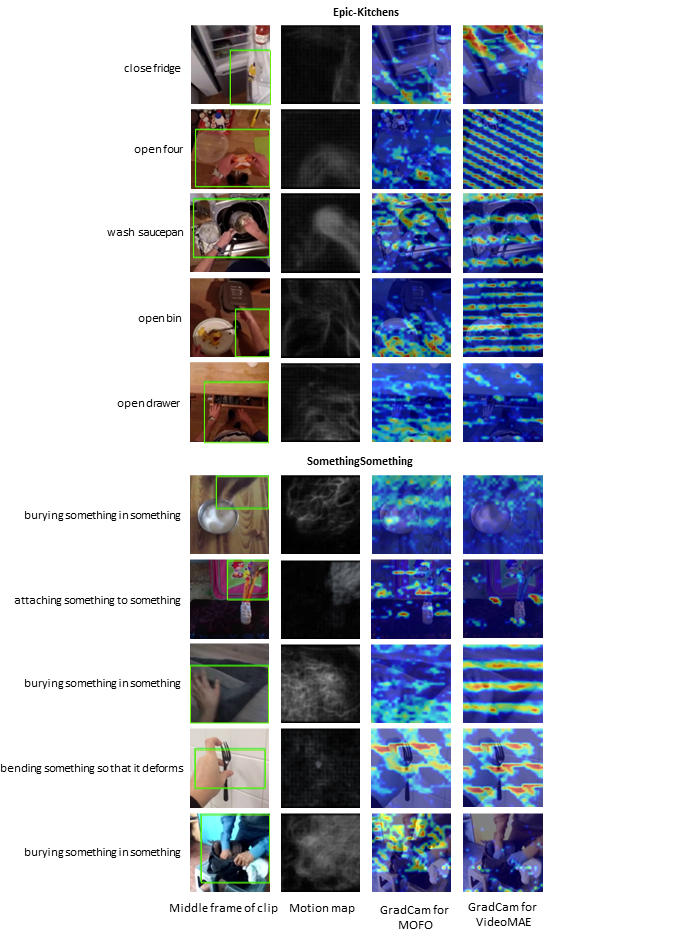}
\caption{We visualise the attention maps generated by GradCAM based on VideoMAE and MOFO for Epic-Kitchens and the Something-Something V2 dataset. The attention maps show that our proposed approach can better capture the motion area.}
\label{fig:CAM}
\end{figure}

\section{ Ablation Study}
We finetune the learned model for action classification to evaluate the learned model as a pretrained
model and train on a new downstream task with the
learned representation. We perform such an evaluation on our
self-supervised model to gain some insights into the generality of the learned features. For finetuning, we follow the same protocol in~\cite{tong2022videomae} to provide a fair comparison and call it regular finetuning. The entire feature encoder and a linear layer are finetuned end-to-end with cross-entropy loss,
The recognition accuracy for our MOFO SSL using regular finetuning is reported in Table \ref{tab:MOFO pretrain} shown as MOFO*. We demonstrate significant performance improvement over the other self-supervised approaches, comparable to the best-supervised approach. All variants of our model are presented in section \ref{ratio_supp} outperformed the existing result using ViT-MAE, but we found that the 75\% inside masking ratio worked the best. Compared to VideoMAE~\cite{tong2022videomae}, our approach achieves significantly better results while the number of backbone parameters remains the same. While MOFO** indicates our result with pretraining on non-motion SSL and MOFO finetuning, which further increases accuracy, MOFO$^\dagger$ denotes the MOFO SSL and MOFO finetuning, which we mentioned in Table~\ref{tab:MOFO pretraining} as MOFO(Proposed), and this provides the greatest performance over the best-performing methods on Epic-Kitchens verb, noun and action classification and on Something Something V2 action classification.

\begin{table}[t!]
\centering
\caption{Human activity recognition on \textbf{Epic-Kitchens} and \textbf{Something-Something V2 (SSV2)} in terms of Top-1 and Top-5 accuracy.
\textcolor{blue}{blue: This is the result computed by us using the public code}
MOFO* is pretrained by our MOFO SSL and uses non-MOFO finetuning.
MOFO** This is our result with pretraining on non-MOFO SSL and has MOFO finetuning.
MOFO$^\dagger$ denotes the MOFO SSL and MOFO finetuning.}
\vspace{2mm}
\resizebox{\linewidth}{!}{
  \begin{tabular}{c|c|c|cc|ccc}
    \toprule
    \multirow{3}{*}{Method} & \multirow{3}{*}{Backbone} & \multirow{3}{*}{Param} & \multicolumn{2}{c|}{\textbf{SSV2}} & \multicolumn{3}{c}{\textbf{Epic-Kitchens}} \\
                        
    & & &\multicolumn{2}{c|}{Action} & \multicolumn{1}{c|}{Verb} & \multicolumn{1}{c|}{Noun} & \multicolumn{1}{c}{Action} \\
    
    &  & & Top-1 & Top-5 &\multicolumn{1}{c|}{Top-1} & \multicolumn{1}{c|}{Top-1} & Top-1 \\ 
        \midrule
        \multicolumn{8}{c}{\textit{Supervised}}\\
        \midrule
            
        $\text{TDN}_{\text{EN}}$~\cite{wang2021tdn} & ResNet101×2  & $88$ & $69.6$ & $92.2$ & -  & -  &-\\
                
        SlowFast~\cite{feichtenhofer2019slowfast} & ResNet101  &$53$& $63.1$ & $87.6$  & $65.6$  & $50.0$  & $38.5$\\
        
        TSM~\cite{lin2019tsm} & ResNet-50  &-& $63.4$ & $88.5$  & $67.9$  & $49.0$  & $38.3$\\

        MViTv1~\cite{fan2021multiscale} & MViTv1-B  & $37$& $67.7$ & $90.9$  &  &-  & -\\
        
        TimeSformer~\cite{bertasius2021space} & ViT-B  & $121$ & $59.9$ &  - & -  &-  & -\\
        
        TimeSformer~\cite{bertasius2021space} & ViT-L  & $430$ & $62.4$ &  - &-  & -  & -\\
   
        ViViT FE~\cite{arnab2021vivit} & ViT-L  & - & $65.9$ & $89.9$ & $66.4$  & $56.8$  &$44.0$\\
        
        Mformer~\cite{patrick2021keeping} & ViT-B  & $109$& $66.5$ & $90.1$& $66.7$  & $56.5$  &$43.1$ \\
        
        Mformer~\cite{patrick2021keeping} & ViT-L  &$382$ & $68.1$ & $91.2$& $67.1$   &$57.6$  & $44.1$\\
   
        Video SWin~\cite{Liu_2022_CVPR} & Swin-B  & $88$& $69.6$ & $92.7$& $67.8$  & $57.0$  & $46.1$\\
        
        \midrule
        \multicolumn{8}{c}{\textit{Self-supervised}}\\
        \midrule
         VIMPAC~\cite{tan2021vimpac} & ViT-L  & $307$ & $68.1$& - & -  & -  & -\\

         BEVT~\cite{wang2022bevt} & Swin-B  & $88$ & $70.6$ & - & -  &-  & -\\
        
    
         VideoMAE~\cite{tong2022videomae} & ViT-B  & $87$ &$70.8$ &$92.4$& \textcolor{blue}{$71.6$}  & \textcolor{blue}{$66.0$}  &\textcolor{blue}{ $53.2$}\\

          ST-MAE~\cite{NEURIPS2022_e97d1081} & ViT-L  & $304$ &$72.1$ & -  & -  & -  &-\\
          
          OmniMAE~\cite{girdhar2022omnimae} & ViT-B  & $87$ &$69.5$ & -& -  & -  & $39.3$\\

       Omnivore(Swin-B)~\cite{girdhar2022omnivore} & ViT-B  & - &$71.4$ & $93.5$& $69.5$  & $61.7$  & $49.9$\\

        \midrule
          \textbf{Ours(MOFO*)} &ViT-B  &$87$& $\bf72.7$ & $\bf94.2$ & $\bf73.0$  & $\bf67.1$ &$\bf54.1$\\
          
          \textbf{Ours(MOFO**)} &ViT-B  &$102 $& $\bf74.7$ & $\bf95.0$ & $\bf74.0$ & $\bf68.0$ &$\bf54.5$\\

          \textbf{Ours(MOFO$^\dagger$)} &ViT-B  &$ 102$& \textbf{75.5} & \textbf{95.3} &  $\bf74.2$   &  $\bf68.1$ & $\bf54.5$ \\
          \bottomrule
        \end{tabular}
      }
        \label{tab:MOFO pretrain}
\end{table}

\section{Domain Generalization}
Domain generalisation aims to build a predictor that can perform well in an unseen test domain, known as out-of-distribution generalisation.
The main objective of this experiment is to learning
video representations that transfer well to a novel previously
unseen dataset. 
We take the MOFO and non-MOFO pretrained models that have already learned features from one dataset and finetune them to adapt them to a new dataset.
Results in Table~\ref{tab:generalization} show that our proposed MOFO model and non-MOFO pretrained model got on-par results;  our MOFO pretrained model's accuracy on SSV2 is marginally higher when pretraining is done on Epic-Kitchens, and marginally worse on Epic-Kitchens when pretraining is done on SSV2.
These results have inspired me to design a self-supervision task to enhance generalisation.

\begin{table}[t!]
\centering
\caption{Human activity recognition on \textbf{Epic-Kitchens} and \textbf{Something-Something V2} in terms of Top-1 accuracy.
\textcolor{blue}{blue: This is the result computed by us using the public code}
MOFO* is pretrained by our MOFO SSL and uses non-MOFO (regular) finetuning.}
\vspace{2mm}
\resizebox{\linewidth}{!}{
  \begin{tabular}{c|c|c|cc|ccc}
    \toprule
    \multirow{3}{*}{Method} & \multirow{3}{*}{Backbone} & \multirow{3}{*}{Pretrain Dataset} & \multicolumn{2}{c|}{\textbf{Something-Something V2}} & \multicolumn{3}{c}{\textbf{Epic-Kitchens}} \\
                        
    & & &\multicolumn{2}{c|}{Action} & \multicolumn{1}{c|}{Verb} & \multicolumn{1}{c|}{Noun} & \multicolumn{1}{c}{Action} \\
    
    &  &  &\multicolumn{2}{c|}{Top-1}  &\multicolumn{1}{c|}{Top-1} & \multicolumn{1}{c|}{Top-1} & Top-1 \\ 

        \midrule
               
         VideoMAE~\cite{tong2022videomae} & ViT-B  & $Something-Something V2$  &\multicolumn{2}{c|}{$70.8$}& \textcolor{blue}{$70.2$}  & \textcolor{blue}{$62.9$}  &\textcolor{blue}{ $50.7$}\\
         VideoMAE~\cite{tong2022videomae} & ViT-B  & $Epic-Kitchens$ &\multicolumn{2}{c|}{\textcolor{blue}{$67.3$}} & \textcolor{blue}{$71.6$}  & \textcolor{blue}{$66.0$}  &\textcolor{blue}{ $53.2$}\\

        \midrule
          \textbf{Ours(MOFO*)} &ViT-B  &$Something-Something V2$& \multicolumn{2}{c|}{$72.7$} & $70.0$  & $62.7$ &$50.6$\\
          
          \textbf{Ours(MOFO*)} &ViT-B  &$Epic-Kitchens $ & \multicolumn{2}{c|}{$67.4$} & $73.0$ & $67.1$ &$54.1$\\

          \bottomrule
        \end{tabular}
      }
        \label{tab:generalization}
\end{table}

\section{Automatic Motion Area Detection}\label{BBox}

\noindent\textbf{Automatic vs.~supervised motion area detection.}
We compare the results using our automatically detected motion areas and the ground truth bounding box annotation provided by~\cite{damen2022rescaling} on the Epic-Kitchens dataset in Table~\ref{fig:bbox_ablation}(b). Our automatic motion detection results are close compared to supervised annotations, as seen in Table~\ref{fig:bbox_ablation}(b), despite the challenging camera motion from the egocentric videos.

\begin{figure}[t!]
    \begin{minipage}[h]{0.5\linewidth}
        \raggedleft

        \includegraphics[width=1.1\textwidth]{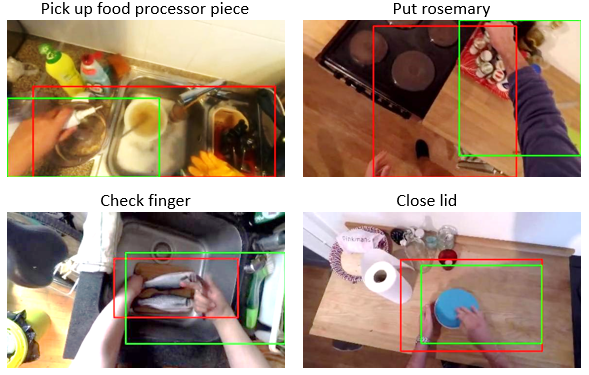}
        \centering
        (a)
    \end{minipage}
    \hspace{0.6cm}
    \begin{minipage}[h]{0.5\linewidth}
    \raggedleft
    \resizebox{1\linewidth}{!}{
    \renewcommand*{\arraystretch}{1.2}
    \begin{tabular}[t]{c|c|c}
        \toprule
        \multirow{3}{*}{Method} & \multirow{3}{*}{Annotation} & \multicolumn{1}{c}{\textbf{Epic-Kitchens}} \\
                        
        & & \multicolumn{1}{c}{Verb}  \\
        &  & Top-1\\ 
        \midrule
        \multirow{2}{*}{MOFO supervision} & Supervised & $73.26$   \\
        & Automatic(ours) & $72.99$ \\
     
        \bottomrule
        \end{tabular}
    }
    \centering
    (b)
    \end{minipage}
    \caption{(a) Comparison between the unsupervised  and supervised motion area detection, \textcolor{green}{green} rectangles indicate the unsupervised while \textcolor{red}{red} ones show supervised detected motion area. (b) Effect of supervised vs. automatic motion area utilisation in MOFO.}
    \label{fig:bbox_ablation}
\end{figure}

We compute the Intersection over the Union (IoU) metric to compare our automatic detector with the supervised annotated bounding boxes on both datasets~\cite{damen2022rescaling,materzynska2020something}. For the Epic-Kitchens dataset, the IoU is $40\%$, and for Something-Something V2, the IoU is $31\%$. Although these numbers are lower, our automatic motion detection only detects motion and ignores unnecessary static objects near the motion. As you can see in Fig.~\ref{fig:bbox_ablation}(a), our automatic motion box still focuses on the area and object of interest, which is the key requirement.

In Fig.~\ref{fig:bbx}, we present additional qualitative examples of our automatic motion area detection compared with the provided supervised annotation for Epic-Kitchens and Something-Something V2 datasets.
These samples show that our proposed automatic motion area detection minimises the impact of the static object in the motion box while highlighting the motion areas.
Our automatic motion box concentrates on the area and item of interest, which is necessary for our proposed approach, even for self-supervision or finetuning.
\begin{figure}[htp!]
\centering
\includegraphics[width=1\linewidth]{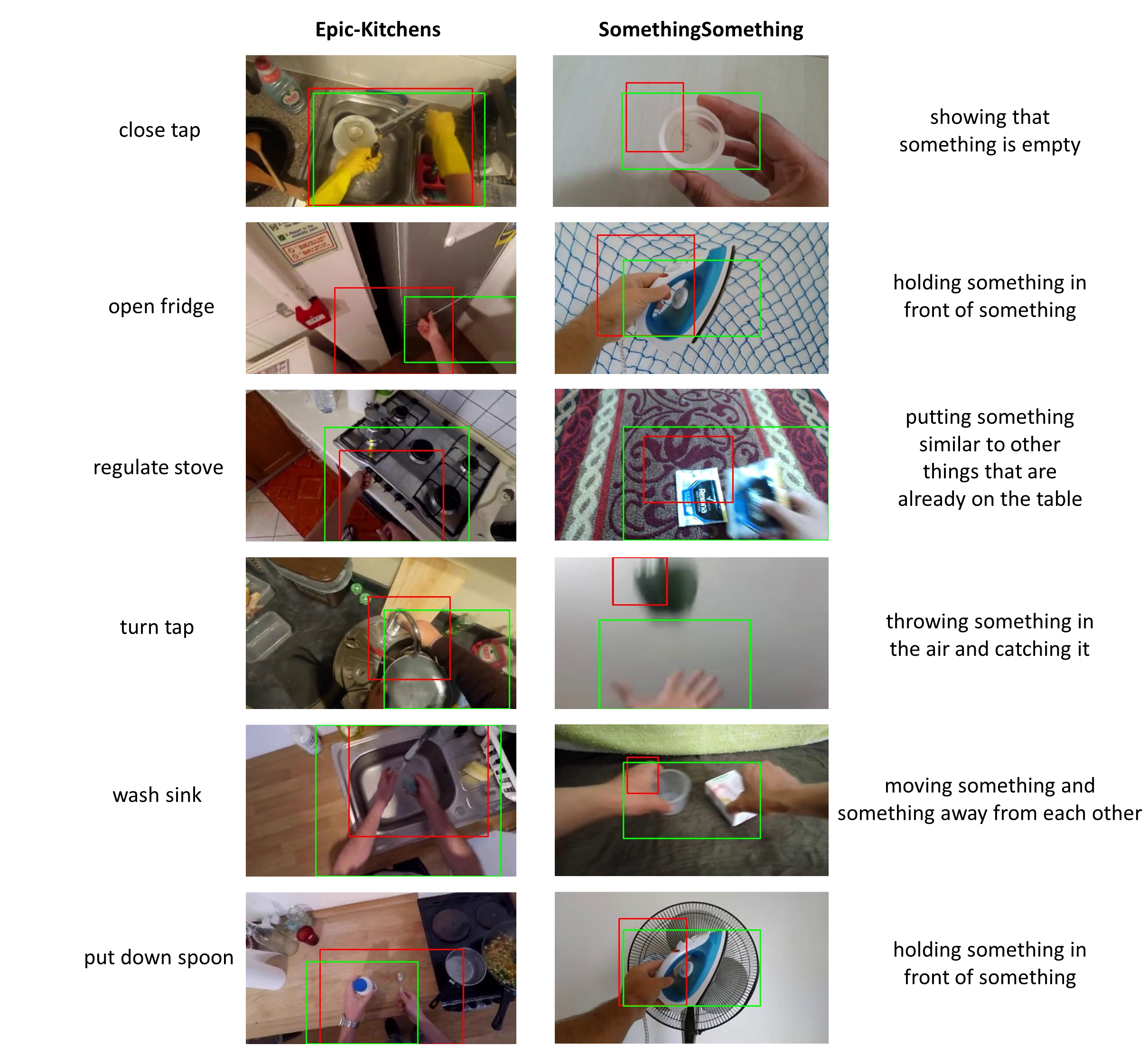}
\caption{Comparison between the unsupervised  and supervised motion area detection, \textcolor{green}{green} rectangles indicate the unsupervised while \textcolor{red}{red} ones show supervised detected motion area.}
\label{fig:bbx}
\end{figure}

\section{Related Work}\label{related}

Self-supervised learning (SSL) is a developing machine learning technique that has the potential to address the issues brought about by over-dependence on labelled data. High-quality labelled data have been essential for many years to develop intelligent systems using machine learning techniques. Consequently, high-quality annotated data costs are a significant bottleneck in the training process.
Grow the research and development of generic AI systems at an inexpensive cost. Self-learning mechanisms with unstructured data are one of the top focuses of AI researchers. Collecting and labelling a wide range of diverse data is almost impossible.
Researchers are developing self-supervised learning (SSL) methods that can pick up on fine details in data to address this issue.
The introduction to self-supervised learning in video understanding is followed by a review of the literature on video action recognition, the downstream task we have recently focused on.

\subsection{Self-supervised video representation learning}\label{ssl works}
The effectiveness of deep learning-based computer vision relies on the availability of a considerable amount of annotated data, which is time-consuming and expensive to obtain. Supervised learning is trained over a given task with a large, manually labelled dataset. In addition to the costly manual labelling, generalisation mistakes and erroneous correlations are other problems with supervised learning.

Large labelled datasets are difficult to create in particular situations, making it challenging to construct computer vision algorithms. Most computer vision applications in the real world use visual categories not included in a common benchmark dataset. In specific applications, visual categories or their appearance are dynamic and vary over time. Therefore, self-supervised learning could be created that uses a limited number of labelled examples to learn to recognise new concepts effectively. A substantial research effort focuses on learning from unlabeled data, which is much easier to acquire in real-world applications. The ultimate goal is to make it possible for machines to comprehend new concepts quickly after only viewing a few labelled instances, similar to how quickly humans can learn.

SSL has gained considerable popularity since its introduction in natural language processing \cite{DevlinCLT19} and computer vision \cite{doersch2015unsupervised,chen2020big,xie2020self} owing to its ability to learn effective data representations without requiring manual labels. 
Acquiring detailed manual labels is arguably more difficult (and often expensive) in many image and video-related tasks, which makes SSL an increasingly popular paradigm in video analysis. 

The goal of video self-supervised learning for computer vision is to learn meaningful video representations without explicit supervision, and the model trains itself to learn one part of the input from another part of the input.
Self-supervised learning algorithms can learn representations by solving pretext tasks that can be formulated using only unlabeled data. These auxiliary tasks can guide the model to learn intermediate representations of data. By solving these tasks, the model learns to extract relevant features from the input data and understand the underlying structural meaning beneficial for practical downstream tasks. Based on the surrogate task employed, the training objective for self-supervised learning is defined, and model parameters are updated through gradient descent to minimise prediction error.
Therefore, models are trained to solve these pretext tasks. As a result, they learn to capture meaningful and useful representations that can be used for various downstream video understanding tasks, such as video action recognition~\ref{AR works}.

Video-based self-supervised learning techniques start from
image tasks. Several specifically designed tasks, including image inpainting~\cite{pathak2016context}, solving jigsaw puzzles~\cite{noroozi2016unsupervised},  and image
colour channel prediction~\cite{zhang2016colorful} are proposed to learn image features. SSL has recently yielded successful results in learning visual representations from unlabeled videos with various pretext tasks~\cite{yun2022patch, caron2021emerging, gupta2022maskvit}. These methods use a backbone that has been pretrained with images or videos in a self-supervised manner to perform tasks on videos, including contrastive learning ~\cite{yun2022patch, guo2022cross, yang2020hierarchical}, self-distillation ~\cite{caron2021emerging}, or Masked Modeling which selects a random section of the input sequence to mask out, and then predicts the features of those sections~\cite{wei2022masked,gupta2022maskvit,tong2022videomae,girdhar2022omnimae}. Many existing works~\cite{fernando2017self,xu2019self,wang2020self} have
been proposed to focus on temporal information, such as
making models sensitive to the temporal differences of input
data.

As mentioned before, earlier works build on a concept of self-supervision by taking RGB frames as input to learning to predict action concepts~\cite{wang2021self}, using Convolutional Neural Networks (CNNs) models to use frame-wise features and average pooling~\cite{karpathy2014large} discarding the temporal order. Thus, frame-wise CNN scores were fed to LSTMs~\cite{donahue2015long} while in two-stream networks~\cite{simonyan2014two}, representations are computed for each RGB frame and every ten stacked optical flow frames. Spatio-temporal 3D CNN filters~\cite{tran2015learning,varol2017long,feichtenhofer2017spatiotemporal, carreira2017quo} model spatio-temporal patterns.Persistence of Appearance, a motion cue proposed by PAN~\cite{zhang2019pan}, allows the network to extract the motion information from adjacent RGB frames directly. 
Vision Transformers (ViTs)~\cite{dosovitskiy2020image,khan2022transformers} have emerged as an effective alternative to traditional CNNs. The architecture of Vision Transformer is inspired by the prominent Transformer encoder~\cite{devlin2018bert,vaswani2017attention} used in natural language processing (NLP) tasks, which process data in the form
of a sequence of vectors or tokens. Like the word tokens in NLP Transformer, ViT generally divides the image into a grid of non-overlapping patches before sending them to a linear projection layer to adjust the token dimensionality. Feed-forward and multi-headed self-attention layers are then used to process these tokens. ViTs have a wide range of applications in numerous tasks due to their capacity to capture global structure through self-attention, such as classification~\cite{zhang2021token,xiong2022m,li2022semmae}, object detection~\cite{chen2022sdae,li2022mvitv2}, segmentation~\cite{choudhury2022guess,caron2021emerging,baldassarre2022towards} and retrieval~\cite{gabeur2020multi}.

Inspired by ViT~\cite{dosovitskiy2020image}, ViViT~\cite{arnab2021vivit} and Timesformer~\cite{bertasius2021space} were the
first two works that successfully implemented a pure transformer
architecture for video classification, improving upon the state of
the art previously set by 3D CNNs.
In these models, the video clip of RGB frames is embedded into 3D patches to produce downsampled feature maps. Then, these encoded 3D patches are encoded by a Video Transformer ~\cite{patrick2021keeping,zhang2022object}. In the following work, ~\cite{arnab2021vivit} defines the tubelet embedding tokenisation method and inspired some other works to represent a video input by extracting non-overlapping, spatiotemporal tubes to propose their method ~\cite{yan2022multiview}.

In another line of research, Masked Autoencoders (MAEs) have recently been demonstrated to be powerful yet conceptually simple and efficient and have proven an effective pretraining paradigm for Transformer models of text~\cite{devlin2018bert}, images~\cite{he2022masked}, and, more recently, videos~\cite{tong2022videomae}.
The learned self-supervised model from the pretext task can be applied to any downstream computer vision tasks, including classification, segmentation, detection, etc. 

Nowadays, encoder-decoder Transformer-based architectures are commonly used in self-supervised learning for video representation learning. These architectures take advantage of the Transformer models' strengths, initially created for natural language processing challenges, and adapt them to process and comprehend video data.
In the context of video representation learning, the encoder-decoder Transformer architecture typically consists of the following components:
\begin{enumerate}

\item \textbf{Encoder}  The encoder processes the input video data and generates a condensed representation of the video. Each video frame or 3D tublets is typically treated as a sequence of features to be input into the Transformer encoder. Multiple layers of self-attentional and feed-forward neural networks can be used in the encoder to capture the video's temporal dependencies, spatial relationships, and long-range dependencies.


\item \textbf{Decoder:} Based on the self-supervised task, the decoder generates a prediction using the encoder's learned representation. The decoder must solve the surrogate task used for self-supervised learning. For instance, if the self-supervised objective is to anticipate the temporal order of shuffled frames, the decoder may correctly predict that order.
\end{enumerate}

In transformer-based architecture, the self-attention mechanism powers both the encoder and decoder. Self-attention architectures typically are made up of a series of transformer blocks. Each transformer block consists of two sublayers: a feed-forward layer and a multi-head self-attention layer. An input is divided into patches,  and attention evaluates each 3D input patch's usefulness before drawing on it to produce the output. The Transformer's self-attention mechanism lets the model focus on different parts of the video frames while considering their dependencies. Therefore, considering their relative importance, it draws from each input component to produce the output.
The query($Q$), key($K$), and value($V$) vectors are the three sets of calculated vectors in the transformer architecture. These are determined by multiplying the input by a linear transformation.


\subsection{Video action recognition}\label{AR works}
\vspace{2mm}
Although it is simple for humans to recognise and categorise actions in video, automating this process is challenging.
 Human action recognition in video is of interest for applications such as automated surveillance~\cite{khan2020human} detecting anomalies in a camera’s field of view that has attracted attention from vision researchers~\cite{vaswani2005shape}, elderly behaviour monitoring~\cite{sarkar2005humanid}, human-computer interaction, content-based video retrieval~\cite{sowmyayani2022stharnet}, and video summarization~\cite{shabani2011improved}. Activity analysis must be able to identify atomic movements like "walking," "bending," and "falling" on their own while monitoring the daily activities of elderly people, for instance~\cite {249}. Therefore, action recognition is a challenging problem with many potential applications.

\paragraph{Action Recognition Datasets}
Human action recognition aims to understand human activities occurring in a video as humans can understand. While some simple actions, like standing, can be recognised from a single frame (image), most human actions are much more complex and occur over a more extended period. Therefore, they must be observed through consecutive frames (video). To assist organisations in understanding real-time action and dynamic, organic movement, AI/ML models use human action datasets. 

\underline{Something-Something V2}~\cite{goyal2017something} This publically available dataset is an extensive collection of human-object interaction of densely labelled 174 video sequences. The dataset was created by many crowd workers performing pre-trained daily human–object interaction physical activities; 220,847 videos and JPG images have variable spatial resolutions and lengths.

Egocentric vision, sometimes known as first-person vision, is a sub-field of computer vision that deals with analysing images and videos captured by a wearable camera, often worn on the head or the chest and thus naturally approximates the wearer's visual field. The idea of using egocentric videos has recently been utilised thanks to novel, lightweight and affordable devices such as GoPro and similars~\cite{nunez2022egocentric}.
As a fundamental problem in egocentric vision, one of the tasks of egocentric action recognition aims to recognise the actions of the camera wearers from egocentric videos. This community did not have an extensive dataset to be used for pertaining or to have a standard dataset for benchmarking until the appearance of the \underline{Epic-Kitchens}~\cite{damen2018scaling,damen2020epic,damen2020rescaling}, the largest and most complete egocentric dataset contains 97 verb classes, 300 noun classes and 3806 action classes. Understanding egocentric videos requires detecting the actor's movement and the object with which the actor interacts.

Several existing methods leveraged object detection to improve egocentric video recognition~\cite{wang2020symbiotic,wang2020symbiotic,wu2019long,ma2016going}, among which~\cite{wu2019long} also incorporate temporal contexts to help understand the ongoing action. These approaches may have limited uses in real-world systems since they demand time-consuming, labour-intensive item detection annotations and are computationally expensive. In contrast, our framework does not depend on costly object detectors. Recently, Shanetal.\cite{shan2020understanding} developed a hand-object detector to locate the active object. When the detector is well-trained, it can be deployed on the target dataset; however, running it on high-resolution frames still costs far more than using our method.

\paragraph{Motion in action recognition:}
Motion cues\cite{akar2022mac,wang2019self,li2021motion} have been recognised as necessary for video understanding in the past few years. Most works use optical flow, a motion representation component in many video recognition techniques, to obtain the statistical motion labels required for their work~\cite{yang2021self}, separating the background from the main objects in optical flow frames.
Optical flow is the pattern of visible motion of objects and edges and helps calculate the motion vector of every pixel in a video frame. Optical flow is widely used in many video processing applications as a motion representation feature that can give important information about the spatial arrangement of the objects viewed and the rate of change of this arrangement. Optical flow-based techniques are sensitive to camera motion since they capture absolute movement.
Optical flow computation is one of the fundamental tasks in computer vision. In practice, the flow has been helpful for a wide range of problems, for example, pose estimation ~\cite{pfister2015flowing}, representation learning~\cite{senturk2022triplednet}, segmentation~\cite{luiten2020unovost}, and even utilised as a tracking substitute for visual signals (RGB images)~\cite{sidenbladh2000stochastic}. Since optical flow can capture continuous or smoothly varying motion, such as motion caused by a change in camera view, it is not a good idea to use it to detect a change in salient objects. To build pixel-level representations from raw high-resolution videos with complex scenes, ~\cite{xiong2021self} proposes a self-supervised representation learning framework based on a flow equivariance objective. This representation is beneficial for object detection.
In another work~\cite{li2019motion}, a multi-task motion-guided video salient object detection network is proposed consisting of two sub-networks. One sub-network is used to detect salient objects in still images, and the other is used to detect motion saliency in optical flow images.
Most motion descriptors use absolute motions and thus only work well when the camera and background are relatively static, such as Fleet \& Jepson's phase-based features~\cite{fleet1993stability} and Viola et al.'s generalised wavelet features~\cite{viola2005detecting}. Therefore, the critical problem is identifying characteristics that accurately capture the motion of hands or objects while impervious to the camera and backdrop motion.

Relying only on optical flow to capture the motion is not a robust solution as it is heavily affected by camera motion.
To mitigate this problem, \cite{wang2019self} presented a self-supervised spatiotemporal video representation by predicting a set of statistical labels derived from motion and appearance statistics using extracting optical flow across each frame and two motion boundaries \cite{dalal2006human} which are obtained by computing gradients separately on the horizontal and vertical components of the optical flow.

In another line of work, masked autoencoder models have been proposed to learn underlying data distribution in a self-supervised manner without explicitly focusing on motion~\cite{tong2022videomae}. Even though this model can perform spatiotemporal reasoning over content, the encoder backbone could be more effective in capturing motion representations. The critical contribution of our work is explicitly imposing motion information in both SSL phases in the self-supervised pretext training without human annotations and then in the finetuning stage, besides introducing an automatic motion detection to detect salient objects and motion in the video without the overhead and limitation of a pretrained and annotated object detector.
\end{alphasection} 
\end{document}